\documentclass{article}
\usepackage{enumerate}
\usepackage{amssymb}
\usepackage{amsmath}
\usepackage{mathrsfs}
\usepackage{graphicx}
\usepackage{subfigure}
\usepackage{color}
\usepackage{amsthm}

\date{}

\usepackage{geometry}
\theoremstyle{remark}

\theoremstyle{definition}

\usepackage{authblk}

\usepackage{mathtools}

\newcommand{\R}{\mathcal{R}}

\newcommand{\E}{\mathcal{E}}
\newcommand{\F}{\mathcal{F}}
\newcommand{\U}{\mathcal{U}}

\usepackage[ruled,vlined]{algorithm2e}

\begin{document}

\title{Full Gradient DQN Reinforcement Learning:\\ A Provably Convergent Scheme
}

\author[1]{K. E. Avrachenkov }
\author[2]{V. S. Borkar}
\author[2]{H. P. Dolhare}
\author[1]{K. Patil}

\affil[1]{INRIA Sophia Antipolis, France 06902\footnote{email: k.avrachenkov@inria.fr, borkar.vs@gmail.com, harshdolhare99@gmail.com, kishor88k@gmail.com.}}
\affil[2]{Indian Institute of Technology, Bombay, India 400076}
\maketitle

\begin{abstract}
 We analyze the DQN reinforcement learning algorithm as a stochastic approximation scheme using the o.d.e.\  (for \textit{`ordinary differential equation'}) approach and point out certain theoretical issues. We then propose a modified scheme called Full Gradient DQN (FG-DQN, for short) that has a sound theoretical basis  and compare it with the original scheme on sample problems. We observe a better performance for FG-DQN.

\medskip


\medskip

\noindent
{\bf AMS(2000) subject classification:} Primary 93E35, Secondary 68T05, 90C40, 93E35
\end{abstract}


\section{Introduction}
Recently we have witnessed tremendous success of Deep Reinforcement Learning algorithms in various application domains.
Just to name a few examples,
DRL has achieved superhuman performance in playing Go \cite{Silver16}, Chess \cite{Silver18} and many Atari video games \cite{Mnih1,Mnih2}. In Chess, DRL algorithms have also beaten the state of the art computer programs, which are based on more or less brute-force enumeration of moves. Moreover, playing Go and Chess, DRL surprised experts with new insights and beautiful strategies \cite{Silver16,Silver18}. We would also like to mention the impressive progress of DRL applications in
robotics \cite{Gu17,Haarnoja18,Peng17}, telecommunications \cite{Luong2019,Qian19,Xiong19} and medicine \cite{Jonsson19,Popova18}.\\

The use of Deep Neural Networks is of course an essential part of DRL. However, there are other paramount elements that contributed to the success of DRL. A starting point for DRL was the Q-learning algorithm of Watkins \cite{Watkins}, which in its original form can suffer from the proverbial curse of dimensionality.
In \cite{Jaakola}, \cite{Tsitsiklis} the convergence of Q-learning has been rigorously established.
Then, in \cite{Gordon96,Gordon99} Gordon has proposed and analyzed fitted Q-learning using a novel architecture based on what he calls `averager' maps.
In \cite{Riedmiller05} Riedmiller has proposed using a neural network for approximating Q-values. There he has also suggested that we treat the right hand side of the dynamic programming equation for Q-values (see equation (\ref{Q-DP}) below) as the `target' to be chased by the left hand side, i.e., the Q-value itself, and then seek to minimize the mean squared error between the two. The right hand side in question also involves the Q-value approximation and \textit{ipso facto} the parameter  itself, which is treated as a `given' for this purpose, as a part of the target, and the minimization is carried out only over the same parameter appearing in the left hand side. This leads to a scheme reminiscent of temporal difference learning, albeit a nonlinear variant of it. The parameter dependence of the target leads to some difficulties because of the permanent shifting of the target itself, what one might call the `dog chasing its own tail' phenomenon.
Already in \cite{Riedmiller05}, frequent instability of the algorithm has been reported.\\

The next big step in improvement of DRL performance was carried out by DeepMind researchers, who elaborated the Deep Q-Network (DQN) scheme \cite{Mnih1}, \cite{Mnih2}.
Firstly, to improve the stability of the algortihm in \cite{Riedmiller05}, they suggested  freezing the parameter value in the target network for several iterates. Thus in DQN, the target network evolves on a slower timescale.
The second successful tweak for DQN has been the use of `experience replay', or averaging over some relevant traces from the past, a notion introduced in \cite{Lin92,Lin}. Then, in \cite{van1,van2}
it was suggested that we introduce a separation of policy estimation and evaluation
to further improve stability.
The latter scheme is called Double DQN.
While various success stories of DQN and Double DQN schemes have been reported, this does not completely fix the theoretical and practical issues.\\

Let us mention that apart from Q-value based methods in DRL, there is another large family of methods based on policy gradient. Each family has its own positive and negative features (for background on RL and DRL methods we recommend the texts
\cite{SuttonBarto,Bertsekas19,Francois-Lavet}). While there has been a notable progress in the theoretical analysis of the policy gradient methods
\cite{Marbach,Sutton99,Bhandari19,Cai19,Agarwal20,Agazzi21}, there are no works establishing convergence of the neural Q-value based methods to the best of our knowledge.\\

In this work, we revisit DQN and scrutinize it as a stochastic approximation algorithm, using the `o.d.e.' (for `ordinary differential equation') approach for its convergence analysis (see \cite{BorkarBook} for a textbook treatment). In fact, we go beyond the basic o.d.e.\ approach to its generalization based on differential inclusions, involving in particular non-smooth analysis. This clarifies the underlying difficulties regarding theoretical guarantees of convergence and also suggests a modification, which we call the Full Gradient DQN, or FG-DQN. We establish theoretical convergence guarantees for FG-DQN and compare it empirically with DQN on sample problems
(forest management \cite{Chades14,Couture16} and cartpole \cite{Barto83,Florian07}), where it gives better performance at the expense of some additional computational overhead per iteration.\\

As was noticed above, another successful tweak for DQN has been the use of `experience replay'. We too incorporate this in our scheme. Many advantages of experience replay have been cited in literature, which we review later in this article. We also unearth an interesting additional advantage of `experience replay' for Bellman error minimization using gradient descent and compare it with the `double sampling' technique of \cite{Baird}. See Sections~\ref{subsec:doublesampling} and \ref{subsec:forestproblem} below.

\section{DQN reinforcement learning}

\subsection{Q-learning}

We begin by recalling the derivation of the original Q-learning scheme \cite{Watkins} to set up the context. Consider a Markov chain $\{X_n\}$ on a finite state space $S := \{1,2,\cdots,s\}$, controlled by a control process $\{U_n\}$ taking values in a finite action space $A=\{1,2,\cdots,a\}$. Its transition probability function is denoted by
$(x, y, u) \in S^2\times A \mapsto p(y|x,u) \in [0,1]$ such that $\sum_y p(y|x,u) = 1 \ \forall \ x, u$. The controlled Markov property then is
$$
P(X_{n+1} = y|X_m, U_m, m \leq n) = p(y|X_n, U_n) \quad \forall\ n \geq 0, \ y \in S.
$$
We call $\{U_n\}$ an admissible control policy. It is called a stationary policy if $U_n = v(X_n) \ \forall n$ for some $v: S \rightarrow A$. A more general notion is that of a stationary randomized policy wherein one chooses the control $U_n$ at time $n$ probabilistically with a conditional law given the $\sigma$-field $\F_n := \sigma(X_m,U_m, m < n; X_n)$ that depends only on $X_n$. That is,
$$
\varphi(u|X_n) := P(U_n = u|\F_n) = P(U_n = u| X_n)
$$
for a prescribed map $x \in S \mapsto \varphi( \cdot | x) \in \mathcal{P}(A) :=$ the simplex of probability vectors on $A$. One identifies such a policy with the map $\varphi$. Denote the set of stationary randomized policies by $\U_{SR}$. In anticipation of the learning schemes we discuss, we impose the `frequent updates' or `sufficient exploration' condition
\begin{equation}
\liminf_{n\uparrow\infty}\frac{1}{n}\sum_{m=0}^{n-1}I\{X_m = x, U_m = u\} > 0 \quad \mbox{a.s.} \quad \forall x, u. \label{freq}
\end{equation}
Given a per stage reward $(x,u)  \mapsto r(x,u)$ and a discount factor $\gamma \in (0, 1)$, the objective is to maximize the infinite horizon expected discounted reward
$$E\left[\sum_{m=0}^\infty \gamma^m r(X_m, U_m)\right].$$
The `value function' $V: S \rightarrow \R$ defined as
\begin{equation}
V(x) = \max E\left[\sum_{m=0}^\infty \gamma^m r(X_m, U_m)\Big| X_0 = x\right], \quad x \in S, \label{val}
\end{equation}
then satisfies the dynamic programming equation
\begin{equation}
V(x) = \max_u \left[r(x,u) + \gamma\sum_y p(y|x,u)V(y)\right],
\quad x \in S. \label{DP}
\end{equation}
Furthermore, the maximizer $v^*(x)$ on the right (chosen arbitrarily if not unique) defines a stationary policy $v^*: S \rightarrow A$ that is optimal, i.e., achieves the maximum in (\ref{val}). Equation (\ref{DP}) is a fixed point equation of the form $V = F(V)$ (which defines the map $F: \R^s \rightarrow \R^s$) and can be solved by the `value iteration' algorithm
\begin{equation}
V_{n+1}(x) = \max_u \left[r(x,u) + \gamma\sum_y p(y|x,u)V_n(y)\right], \quad n \geq 0, \label{ValIter}
\end{equation}
beginning with any $V_0 \in \R^s$. $F$ can be shown to satisfy
$$\|F(x) - F(y)\|_\infty \leq \gamma\|x - y\|_\infty,$$
i.e., it is an $\| \cdot \|_\infty$-norm contraction. Then (\ref{ValIter}) is a standard fixed point iteration of a contraction map and converges exponentially to its unique fixed point $V$.\\

Now define Q-values as the expression in square brackets in (\ref{DP}), i.e.,
$$Q(x,u) = r(x,u) + \gamma\sum_y p(y|x,u)V(y), \quad x \in S, \ u \in A.$$
If the function $Q( \cdot , \cdot )$ is known, then the optimal control at state $x$ is found by simply maximizing
$Q(x, \cdot )$ without requiring the knowledge of reward or transition probabilities. This makes it suitable for data-driven algorithms of reinforcement learning. By (\ref{DP}),  $V(x) = \max_u Q(x,u)$. The Q-values then satisfy their own dynamic programming equation
\begin{equation}
Q(x,u) = r(x,u) + \gamma\sum_y p(y|x,u) \max_v Q(y,v), \label{Q-DP}
\end{equation}
which in turn can be solved by the `Q-value iteration'
\begin{equation}
Q_{n+1}(x,u) = r(x,u) + \gamma\sum_y p(y|x,u) \max_v Q_n(y,v), \quad x \in S, \ u \in A. \label{QIter}
\end{equation}
What we have gained at the expense of increased dimensionality is that the nonlinearity is now inside the conditional expectation w.r.t.\ the transition probability function. This facilitates a stochastic approximation algorithm \cite{BorkarBook} where we first replace this conditional expectation by actual evaluation at a real or simulated random variable $\zeta_{n+1}(x,u)$ with law $p( \cdot | x,u)$, and then make an incremental correction to the current guess based on it. That is, replace (\ref{QIter}) by
\begin{equation}
Q_{n+1}(x,u) = (1 - a(n))Q_n(x,u) + a(n)\left(r(x,u) + \gamma\max_v Q_n(\zeta_{n+1}(x,v),v)\right) \label{QIter2}
\end{equation}
for some $a(n) > 0$. The Q-learning algorithm does so using a single run of a real or simulated controlled Markov chain $(X_n, U_n), n \geq 0,$ so that:
\begin{itemize}
\item at each time instant $n$, $(X_n, U_n)$ are observed and the $(X_n, U_n)$th component of $Q$ is updated, leaving other components of $Q_n(\cdot, \cdot)$ unchanged,

\item this update follows (\ref{QIter2}) where $\zeta_{n+1}(x,u)$ with $x = X_n, u = U_n$, gets replaced by $X_{n+1}$, which indeed has the conditional law $p(\cdot | X_n, U_n)$ as required,

\item $\{a(n)\}$ are positive scalars in $(0, 1)$ chosen to satisfy the standard Robbins-Monro conditions of stochastic approximation \cite{BorkarBook}, i.e.,
\begin{equation}
\sum_na(n) = \infty, \quad \sum_na(n)^2 < \infty. \label{RM}
\end{equation}
\end{itemize}
It is more convenient to write the resulting Q-learning algorithm as
\begin{eqnarray}
Q_{n+1}(x,u) &=& Q_n(x,u) + a(n)I\{X_n = x, U_n = u\}\Bigg(r(x,u) + \nonumber \\
&& \gamma \max_v Q_n(X_{n+1}, v) -Q_n(x,u)\Bigg) \quad \forall \ x,u, \label{QLearn}
\end{eqnarray}
where $I\{ \cdots \} :=$ the indicator  random variable that equals $1$ if `$\cdots$' holds and $0$ if not. The fact that only one component is being updated at a time makes this an asynchronous stochastic approximation. Nevertheless, it exhibits the well known `averaging effect' of stochastic approximation whereby it is a data-driven scheme that emulates (\ref{QIter}) and exhibits convergence a.s.\ to the same limit, viz., $Q$. For formal proofs, see \cite{Jaakola,Tsitsiklis,WatkinsDayan}.\\

\subsection{DQN learning}

The raw Q-learning scheme (\ref{QLearn}), however, does inherit the `curse of dimensionality' of MDPs. One common fix is to replace $Q$ by a parametrized family $(x,u,\theta) \mapsto Q(x,u;\theta)$ (where we again use the notation $Q(\cdot, \cdot; \cdot)$ by abuse of terminology so as to match standard usage). Here $\theta \in \Theta \subset \R^d$ for a moderate $d \geq 1$ and the objective is to learn the `optimal' approximation $Q(\cdot, \cdot; \theta^*)$ by iterating in $\Theta$. For simplicity, we take $\Theta = \R^d$. One natural performance measure is the `DQN Bellman error'
\begin{equation}
\E(\theta) := E\left[\left(Z_n - Q(X_n,U_n;\theta)\right)^2\right], \label{Bellman}
\end{equation}
where
$$
Z_n := r(X_n, U_n) + \gamma\max_v Q(X_{n+1},v; \theta_n)
$$
is the `target' that is taken as a {\it given quantity} and the expectation is w.r.t.\ the stationary law of $(X_n,U_n)$. For later reference, note that this is different from the `true Bellman error'
\begin{equation}
\bar{\E}(\theta) := E\left[\left(r(X_n, U_n) + \gamma\sum\nolimits_y p(y|X_n,U_n)
\max_v Q(y,v; \theta) - Q(X_n,U_n;\theta)\ \right)^2\right]. \label{Ebar}
\end{equation}

%

The stochastic gradient type scheme based on the empirical semi-gradient of $\E(\cdot)$ then becomes
\begin{equation}
\theta_{n+1} = \theta_n +a(n)(Z_n - Q(X_n, U_n;\theta_n))\nabla_\theta Q(X_n, U_n; \theta_n), \quad n \geq 0. \label{DQN}
\end{equation}\\

\subsection{Experience replay}

An important modification of the DQN scheme has been the incorporation of `experience replay'. The idea is to replace the term multiplying $a(n)$ on the right hand side of (\ref{DQN}) by an empirical average over traces of transitions from past that are stored in memory. The algorithm then becomes
\begin{eqnarray}
\lefteqn{\theta_{n+1} = \theta_n + \frac{a(n)}{M}\times} \nonumber \\
&&\sum_{m=1}^M\Bigg((Z_{n(m)} - Q(X_{n(m)}, U_{n(m)}))\nabla_\theta Q(X_{n(m)}, U_{n(m)}; \theta_{n(m)})\Bigg), \ n \geq 0, \label{DQN2}
\end{eqnarray}
where $(X_{n(m)}, U_{n(m)}), 1 \leq m \leq N,$ are samples from past. This has multiple advantages. Some that have been cited in literature are as follows.
\begin{enumerate}

\item As in the mini-batch stochastic gradient descent for empirical risk  minimization in machine learning, it helps reduce variance. It also diminishes effects of anomalous transitions.

\item Training based on only the immediate experiences ($\approx$ samples) tends to overfit the model to current data. This is prevented by experience replay. In particular, if past samples are randomly picked, they are less correlated.

\item The re-use of data leads to data efficiency.

\item Experience replay is better suited for delayed rewards or costs, e.g., when the latter are realized only at the end of a long episode or epoch.

\end{enumerate}

There are also variants of basic experience replay, e.g., \cite{schaul}, which replaces purely random sampling from past by  a non-uniform sampling which picks a sample with probability proportional to its absolute Bellman error.\\

We shall be implementing experience replay a little differently in the variant we describe next, which has yet another major advantage from a theoretical standpoint in the specific context of our scheme.\\

\subsection{Double DQN learning}

One more modification of the vanilla DQN scheme is doing the policy selection according the local network \cite{van1,van2}. The target
network is still used in $Z_n$ and is updated on a slower
time scale. The latter can be represented with another
set of parameters $\bar{\theta}_n$. Thus, the iterate for
the Double DQN scheme can be written as follows:
\begin{equation}
\theta_{n+1} = \theta_n +a(n)(Z_n - Q(X_n, U_n; \theta_n))\nabla_\theta Q(X_n, U_n; \theta_n), \quad n \geq 0, \label{DDQN}
\end{equation}
with
$$
Z_n := r(X_n, U_n) + \gamma
Q(X_{n+1},v; \bar{\theta}_n)\Big|_{v = \mbox{argmax}_{v'} Q(X_{n+1}, v' ; \theta_n)}.
$$
For the sake of comparison, in the vanilla DQN one has:
$$
Z_n := r(X_n, U_n) + \gamma
Q(X_{n+1},v; \bar{\theta}_n)\Big|_{v = \mbox{argmax}_{v'} Q(X_{n+1}, v' ; \bar{\theta}_n)}.
$$
Note that in Double DQN, the selection and evaluation of the policy is done separately.
According to \cite{van1,van2} this modification improves the stability of the DQN learning. One can also combine Double DQN with
experience replay \cite{van2}.\\

\section{The issues with DQN learning}

The expression for DQN learning scheme is appealing because of its apparent similarity with  the very successful temporal difference learning for policy evaluation, not to mention its empirical successes, including some high profile ones such as \cite{Mnih2}. Nevertheless, a good theoretical justification seems lacking. The difficulty arises from the fact that the `target' $Z_n$ is not something extraneous, but is also a function of the operative parameter $\theta_n$. In fact, this becomes apparent once we expand $Z_n$ in (\ref{DQN}) to write
\begin{eqnarray}
\theta_{n+1} &=& \theta_n +a(n)(r(X_n,U_n) + \gamma\max_v Q(X_{n+1}, v; \theta_n) - Q(X_n, U_n;\theta_n))\times \nonumber \\
&& \nabla_\theta Q(X_n, U_n; \theta_n), \ n \geq 0. \label{DQN3}
\end{eqnarray}
Write
\begin{equation}
\tilde{\E}(\theta, \bar{\theta}) := E\left[\left(r(X_n, U_n) + \gamma\max_v Q(X_{n+1},v; \bar{\theta}) - Q(X_n,U_n;\theta) \right)^2\right],
\end{equation}
where $E[ \cdot ]$ is the stationary expectation as before. Consider the `off-policy' case, i.e., $\{(X_n, U_n)\}$ is the state-action sequence of a controlled Markov chain satisfying (\ref{freq}) with a pre-specified stationary randomized policy that does not depend on the iterates. (As we point out later, the `on-policy' version, which allows for the latter adaptation, has additional issues.) If we apply the `o.d.e.\ approach' for analysis of stochastic approximation  (see, e.g., \cite{BorkarBook} for a textbook treatment), we get the limiting o.d.e.\ as
$$\dot{\theta}(t) = -\nabla_1\tilde{\E}(\theta(t), \theta(t)),$$
where $\nabla_i$ denotes gradient with respect to the $i$th argument of $\tilde{\E}(\cdot, \cdot)$ for $i = 1,2$. Thus it is a partial stochastic gradient descent wherein only the gradient with respect to the first occurrence of the variable is used. Unlike gradient dynamics, there is no reason why such dynamics should converge. It was already mentioned that in case of linear function approximation, the DQN iteration bears a similarity with TD(0), except for the nonlinear `max' term. The o.d.e.\ proof of convergence for TD(0) does not carry over to DQN precisely because the stochastic approximation version leads to the interchange of the conditional expectation and max operators. The other issue is that in TD(0), the linear operator in question is a contraction w.r.t.\ the weighted $L_2$-norm weighted by the stationary distribution. That argument also fails for DQN because of presence of the max operator.\\

That said, there is already a tweak that treats the first occurrence of $\theta$ on the RHS, i.e., that inside the maximizer, as the `target' being followed, and updates it only after several (say, $K$) iterates. In principle, this implies a delay in the corresponding input to the iteration and with decreasing stepsizes, introduces only an asymptotically negligible additional error, so that the limiting o.d.e.\ remains the same (\cite{BorkarBook}, Chapter 6). This is also the case for Double DQN.\\

Suppose on the other hand that in DQN or Double DQN we consider  a small constant stepsize $a(n) \equiv a > 0$ and let $K$ be large, so that with a fixed target value, the algorithm nearly minimizes the Bellman error before the target is updated. Then, assuming the simpler `off-policy' case again, the limiting o.d.e.\ \textit{for the target}, treating the multiple iterates between its successive iterates as a subroutine, is
\begin{equation}
\dot{\theta}(t) = -\nabla_1\tilde{\E}(x, \theta(t))\Big|_{x = \mbox{argmax}(\tilde{\E}( \cdot , \theta(t)))}. \label{chase}
\end{equation}
There is no obvious reason why this  should converge either. In fact the right hand side would be $\approx$ the zero vector near the current maximizer and the evolution of the o.d.e.\ and the iteration would be very slow. Of course, this is a limiting case of academic interest only, stated to underscore the  fact that it is difficult to get convergent dynamics out of the DQN learning scheme. This motivates our modification, which we state in the next subsection.\\

\section{Full Gradient DQN}

We propose the obvious, viz., to treat both occurrences of the variable $\theta$ on equal footing, i.e., treat it as a single variable, and then take the full gradient with respect to it. The iteration now is
\begin{eqnarray}
\theta_{n+1} &=& \theta_n - a(n)\left(r(X_n,U_n) + \gamma\max_v Q(X_{n+1}, v; \theta_n) - Q(X_n, U_n;\theta_n)\right)\times \nonumber \\
&& \left(\gamma\nabla_\theta Q(X_{n+1}, v_n; \theta_n) - \nabla_\theta Q(X_n, U_n; \theta_n)\right)
\label{FG-DQN}
\end{eqnarray}
for $n \geq 0$, where  $v_n \in \mbox{Argmax} Q(X_{n+1}, \cdot ; \theta_n)$ chosen according to some tie-breaking rule when necessary. Note that when the maximizer in the term involving the max operator is not unique, one may lose its differentiability, but the expression above still makes sense in terms of the Frechet sub-differential, see Appendix. We assume throughout that $\{X_n\}$ is a Markov chain controlled by the control process $\{U_n\}$ generated according to a fixed stationary randomized policy $\varphi \in \U_{SR}$. Other simulation scenarios are possible for the off-policy set-up. For example, we  may replace the triplets $(X_n, U_n, X_{n+1})$ on the right hand side by triplets $(X_n', U_n', Y_n')$ where $\{X_n'\}$ are generated i.i.d.\ according to some distribution with full support and $(U_n', Y_n')$ are generated with conditional law $P(U_n' = u, Y_n' = y| X_n' = x) = \varphi(u|x)p(y|x,u)$, conditionally independent of all other random variables generated till $n$ given $X_n'$. The analysis will be similar. Yet another possibility is that of going through the relevant pairs $(x,u)$ in a round robin fashion.\\

We modify (\ref{FG-DQN}) further by replacing the right hand side  as follows:
\begin{eqnarray}
\theta_{n+1} &=& \theta_n - a(n)\Bigg(\overline{(r(X_n,U_n) + \gamma\max_v Q(X_{n+1}, v; \theta_n) - Q(X_n, U_n;\theta_n))}\times \nonumber \\
&& \left(\gamma\nabla_\theta Q(X_{n+1}, v_n; \theta_n) - \nabla_\theta Q(X_n, U_n; \theta_n)\right)  + \xi_{n+1}\Bigg)\nonumber \\
\ && \ \label{FG-DQN_expreplay}
\end{eqnarray}
for $n \geq 0$, where $\{\xi_n\}$  is extraneous i.i.d.\  noise componentwise distributed independently and uniformly on $[-1,1]$, and the overline stands for a modified form of experience replay which comprises of averaging at time $n$ over past traces sampled from $(X_k, U_k, X_{k+1}), k \leq  n$, for which $X_k = X_n, U_k = U_n$.  We analyze the asymptotic behavior of this scheme in the remainder of this section in the `off-policy' case, i.e., we use a prescribed stationary randomized policy $\varphi \in \U_{SR}$.\\

We make the following key assumptions:\\

\noindent \textbf{(C1)} (Assumptions regarding the function $Q( \cdot , \cdot ; \cdot)$)
\begin{enumerate}
\item  The map $(x,u;\theta) \mapsto Q(x,u;\theta)$ is bounded and twice continuously differentiable in $\theta$  with bounded first and second derivatives;

\item For each choice of $x \in S$, the set of $\theta$ for which the maximizer of $Q(x, \cdot; \theta)$ is not unique, is the complement of an open and dense set and has Lebesgue measure zero;

\item Call $\hat{\theta}$ a critical point of $\E(\cdot)$ (which is defined in terms of $Q$) if the zero vector is contained in the (Frechet) subdifferential $\partial^-\E(\hat{\theta})$ (see the Appendix for a definition). We assume that there are at most finitely many such points.
\end{enumerate}

\bigskip

We also assume:

\bigskip

\noindent \textbf{(C2)} (Stability assumption)\\

 The iterates remain a.s.\ bounded, i.e.,
\begin{equation}
\sup_n\|\theta_n\| < \infty \ \mbox{a.s.} \label{stability}
\end{equation}

\bigskip

Our final assumption is a bit more technical. Rewrite the term
$$
\overline{(r(X_n,U_n) + \gamma\max_v Q(X_{n+1}, v; \theta_n) - Q(X_n, U_n;\theta_n))}
$$
as
$$
\sum_y p(y|X_n,U_n)\left(r(X_n,U_n) + \gamma\max_v Q(y, v; \theta_n) - Q(X_n, U_n;\theta_n)\right)$$
$$+ \ \varepsilon(X_n,U_n,\theta_n)$$
where the error term $\varepsilon(X_n,U_n,\theta_n)$ captures the difference between the empirical conditional expectation using experience replay and the actual conditional expectation. We assume that:\\

\noindent \textbf{(C3)} (Assumption regarding the residual error in experience replay)\\

The error terms $\{\varepsilon(X_n,U_n,\theta_n)\}$ satisfy
$$
\varepsilon(X_n,U_n,\theta_n) \to 0 \ \mbox{a.s.\ and} \ \sum_na(n)E[|\varepsilon(X_n,U_n,\theta)|] |_{\theta = \theta_n} < \infty \ \mbox{a.s.},
$$
where the expectation is taken w.r.t. the stationary distribution of the state-action pairs.

\bigskip

We comment on these assumptions later. Recall the true Bellman error $\bar{\E}( \cdot )$ defined in (\ref{Ebar}).

\bigskip

\noindent \textbf{Theorem 1} The sequence $\{\theta_n\}$ generated by FG-DQN converges a.s.\ to a sample path dependent critical point of $\bar{\E}(\cdot)$.\\

\noindent \textbf{Proof:} For notational ease, write
$$\epsilon(n) := -\varepsilon(X_n,U_n,\theta_n)\left(\gamma\nabla_\theta Q(X_{n+1}, v_n; \theta_n) - \nabla_\theta Q(X_n, U_n; \theta_n)\right),$$
where $v_n$ is chosen from Argmax $Q(X_{n+1}, \cdot ; \theta_n)$ as described earlier. Consider the iteration
\begin{eqnarray}
\theta_{n+1} &=& \theta_n - a(n)\times \nonumber \\
&&\Bigg(\Big(\sum_y p(y|X_n,U_n)(r(X_n,U_n) + \gamma\max_vQ(y, v; \theta_n)
- Q(X_n, U_n;\theta_n))\Big)\times \nonumber \\
&& \left(\gamma\nabla_\theta Q(X_{n+1}, v_n; \theta_n) - \nabla_\theta Q(X_n, U_n; \theta_n)\right) + \epsilon(n) + \xi_{n+1}\Bigg)\nonumber \\
\ && \ \label{FG-DQN2}
\end{eqnarray}
for $n \geq 0$.
Adding and subtracting the one step conditional expectation of the RHS with respect to
$\F_n' := \sigma(X_m,U_m, m \leq n)$, we have
\begin{eqnarray}
\theta_{n+1} &=& \theta_n - a(n)\times \nonumber \\
&&\left(\sum_{y} p(y|X_n, U_n)(r(X_n,U_n) + \gamma\max_v Q(y, v; \theta_n)
- Q(X_n, U_n;\theta_n))\right) \nonumber \\
&& \times\left(\sum_{y} p(y|X_n,U_n)\left(\gamma\nabla_\theta Q(y, u_n(y); \theta_n) - \nabla_\theta Q(X_n, U_n; \theta_n)\right)\right) \nonumber \\
&&  + \ a(n)\epsilon(n) + \ a(n)M_{n+1}(\theta_n)  \ \label{FG-DQN3}
\end{eqnarray}
where $u_n(y) \in$ Argmax $Q(y, \cdot; \theta_n)$ is chosen as described earlier, and $\{M_n(\theta_{n-1})\}$ is a martingale difference sequence w.r.t.\ the sigma fields $\{\F_n'\}$, given by
\begin{eqnarray*}
\lefteqn{M_{n+1}(\theta_n) =} \\
&& \Bigg(\Big(\sum_y p(y|X_n,U_n)(r(X_n,U_n) + \gamma\max_v Q(y, v; \theta_n)
- Q(X_n, U_n;\theta_n))\Big)  \\
&& \times\left(\gamma\nabla_\theta Q(X_{n+1}, v_n; \theta_n) - \sum_y p(y|X_n,U_n)\gamma\nabla_\theta Q(y, u_n(y); \theta_n)\right)   + \xi_{n+1}\Bigg).
\end{eqnarray*}
Because of our assumptions on $Q(\cdot,\cdot;\cdot)$ and $\{\xi_n\}$, $M_n(\cdot)$ will have derivatives uniformly bounded in $n$ and therefore a uniform linear growth w.r.t.\ $\theta$. The same holds for the expression multiplying $a(n)$ in the first term on the right. We shall analyze this iteration as a stochastic approximation with Markov noise $(X_n,U_n), n \geq 0,$ and martingale difference noise $M_{n+1}, n \geq 0$ (\cite{BorkarBook}, Chapter 6).\\

The difficult terms  are those of the form $\gamma\nabla_\theta Q(y, u; \theta)$ above, because all we can say about them is that :
\begin{eqnarray*}
\lefteqn{\nabla_\theta Q(y, u; \theta) \ \in \ G(y, \theta) :=} \\
&& \left\{\sum_v\psi(v|y)\nabla_\theta Q(y, v; \theta) : \psi(\cdot|y) \in \ \mbox{Argmax}_{\phi(\cdot|y)}\left(\sum_u\phi(u|y) Q(y, u; \theta)\right)\right\}.
\end{eqnarray*}
Define correspondingly the set-valued map
$$(x, u, \theta) \mapsto H(x, u, \theta)$$
by
$$
H(x, u; \theta) := \overline{co}\Bigg(\Bigg\{\Big(\sum_{y} p(y|x, u)(r(x,u)
+ \gamma\max_v Q(y, v; \theta) - Q(x, u;\theta))\Big)
$$
$$
\times\sum_{y} p(y|x,u) \left(\gamma\nabla_\theta Q(y, v_j; \theta)
- \nabla_\theta Q(x, u; \theta)\right) : v_j \in \ \mbox{Argmax} Q(y, \cdot; \theta) \Bigg\}\Bigg)
$$
$$
=  \Bigg\{\Big(\sum_{y} p(y|x, u)(r(x,u) + \gamma\max_v Q(y, v; \theta) - Q(x, u;\theta))\Big)
$$
$$
\times\sum_{y} p(y|x,u)\left(\gamma\nabla_\theta \bar{Q}(y, \pi_y; \theta)
- \nabla_\theta Q(x, u; \theta)\right) : \pi_y
 \in \ \mbox{Argmax}
\bar{Q}(y, \cdot; \theta) \Bigg\}
$$
where $\bar{Q}(y,\psi;\theta) := \sum_u \psi(u|y)Q(y,u;\theta)$ for $\psi \in \U_{SR}$. Then (\ref{FG-DQN3}) can be written
 in the more convenient form as the stochastic recursive inclusion (\cite{BorkarBook}, Chapter 5) given by
\begin{equation}
\theta_{n+1} \ \in \ \theta_n - a(n)\Bigg(H(X_n, U_n; \theta_n) + \epsilon(n) + M_{n+1}(\theta_n)\Bigg). \label{short}
\end{equation}
We shall now use Theorem 7.1 of \cite{Yaji}, pp.\ 355, for which we need to verify the assumptions (A1)-(A5), pp.\ 331-2, therein. We do this next.
\begin{itemize}

\item (A1) requires $H(y,\phi,\theta)$ to be nonempty convex compact valued and upper semicontinuous, which is easily verified. It is also bounded by our assumptions on $Q(\cdot, \cdot; \cdot)$.

\item $S_n$ of \cite{Yaji} corresponds to our $(X_n, U_n)$ and (A2) can be verified easily.

\item (A3) are the standard conditions on $\{a(n)\}$ also used here.

\item $M_{n+1}(\theta_n), n \geq 0$, defined above, has linear growth in $\|\theta_n\|$ as observed above. Thus (\ref{stability}) implies that for some $K \in (0,\infty)$,
$$\sum_{m=0}^na(m)^2E\left[\|M_{m+1}(\theta_m)\|^2|\F_m\right] \leq K(1 + \sup_m\|\theta_m\|^2)\sum_ma(n)^2 < \infty \ \mbox{a.s.}$$
This implies that $\sum_{m=0}^{n-1}a(m)M_{m+1}(\theta_m)$ is an a.s.\ convergent martingale by Theorem 3.3.4, pp.\ 53-4, \cite{BorkarProb}.  This verifies (A4).

\item (A5) is the same as (\ref{stability}) above.
\end{itemize}
Let $\mu(x,u) :=$ the stationary probability $P(X_n = x, U_n = u)$ under $\varphi$. Then Theorem 7.1 of \cite{Yaji} applies and allows us to conclude that the iterates will track the asymptotic behavior of the differential inclusion
\begin{equation}
    \dot{\theta}(t) \in -\sum_{x,u} \mu(x,u) H(x, u, \theta(t)).
    \label{incl}
    \end{equation}
Now we make the important observation that under our hypotheses on the function $Q(\cdot , \cdot ; \cdot)$ (see 2.\ of (C1)), for all $x,u$ and Lebesgue-a.e.\ $\theta$ belonging to some open dense set $O$, $H(x,u,\theta)$ is the singleton corresponding to Argmax $Q(x, \cdot ; \theta) = \{u\}$ for some $u \in A$. Furthermore, in this case, the RHS of (\ref{incl}) reduces to $-\nabla\E(\theta(t))$. Since $\{\xi_n\}$ has density w.r.t.\ the Lebesgue measure, so will $\{\theta_n\}$ and therefore by (C1), $\theta_n \in O \ \forall n$, a.s. Let
$$
L(x,u; \theta) := \frac{1}{2}\left(r(x,u) + \gamma\sum_y p(y|x,u)\max_v Q(y,v; \theta)
- Q(x,u; \theta)\right)^2
$$
denote the instantaneous Bellman error. Then
$$
\bar{\E}(\theta) = \sum_{x,u} \mu(x,u) L(x,u;\theta).
$$
Write $\hat{\E}(\theta')$ for  $\bar{\E}(\theta)$ evaluated at a possibly random $\theta'$, in order to emphasize the fact that while $\bar{\E}(\cdot)$ is defined in terms of an expectation, a random argument of $\hat{\E}(\cdot)$ is not being averaged over. We use an analogous notation for other quantities in what follows.  Applying the Taylor formula to
$\bar{\E}(\cdot)$, we have,
$$\hat{\E}(\theta_{n+1}) = \hat{\E}(\theta_n) + \sum_{x,u}\mu(x,u)\langle\nabla_\theta L(x,u;\theta), \theta_{n+1} - \theta_n\rangle + O(a(n)^2).$$
But by (\ref{FG-DQN3}), a.s.,
\begin{eqnarray*}
\theta_{n+1} - \theta_n &=& a(n)\Big(-\nabla_\theta L(X_n,U_n;\theta_n) + \epsilon(n) + M_{n+1}(\theta_n)\Bigg) \\
&=&  a(n)\Bigg(-\sum_{x,u}\mu(x,u)\nabla_\theta L(x,u;\theta_n) + \epsilon(n) + \widetilde{M}_{n+1}(\theta_n) + O(a(n)^2)\Bigg),
\end{eqnarray*}
where we have replaced $\nabla_\theta L(X_n,U_n;\theta_n)$ with $\sum_{i,u}\mu(i,u)\nabla_\theta L(i,u;\theta_n)$, i.e., with the state-action process $(X_n,Z_n)$ averaged w.r.t.\ its stationary distribution (recall that under our randomized stationary Markov policy, it is a Markov chain). This uses a standard (though lengthy) argument for stochastic approximation with Markov noise that converts it to a stochastic approximation with martingale difference noise using the associated parametrized Poisson equation, at the expense of: $(i)$ adding an additional martingale difference noise term that we have added to $M_{n+1}(\theta_n)$ to obtain the combined martingale difference noise $\widetilde{M}_{n+1}(\theta_n)$, and, $(ii)$ another $O(a(n)^2)$ term that comes from the difference of the solution of the Poisson equation evaluated at $\theta_n$ and $\theta_{n+1}$, which is $O(\|\theta_{n+1} - \theta_n\|) = O(a(n))$, multiplied further by an additional $a(n)$ from (\ref{FG-DQN3}) to give a net error that is $O(a(n)^2)$. See \cite{Ben} for a classical treatment of this passage.

Hence for suitable constants $0 < K_1, K_1' < \infty$,
\begin{eqnarray}
E[\hat{\E}(\theta_{n+1})|\F'_n] &\leq& \hat{\E}(\theta_n) + a(n)\Bigg(
-\| \sum_{x,u} \mu(x,u)\nabla_\theta L(x,u;\theta_n)\|^2 \nonumber \\
&& + K_1\sum_{x,u}\mu(x,u)|\varepsilon(x,u,\theta_n)| + K_2 a(n)^2\Bigg) \nonumber \\
&\leq& \hat{\E}(\theta_n) + a(n)\Bigg( K_1 \sum_{x,u}\mu(x,u)|\varepsilon(x,u,\theta_n)| + K_2a(n)^2\Bigg),  \label{almost}
\end{eqnarray}
where we have used (C1). In view of (C3) and the fact $\sum_na(n)^2 < \infty$, the `almost supermartingale' convergence theorem (Theorem 3.3.6, p.\ 54, \cite{BorkarProb}) implies that  $\hat{\E}(\theta_n)$ converges a.s.   This is possible only if
\begin{eqnarray*}
\theta_n &\to& \left\{\theta :  \mbox{the zero vector is in} \ \sum_{x,u}\mu(x,u)H(x,u;\theta)\right\} \\
&=& \left\{\theta : \theta  \ \mbox{is a critical point of} \ \sum_{x,u}\mu(x,u)H(x,u;\theta)\right\}.
\end{eqnarray*}

\noindent By property (P4) of the Appendix, it follows that $H(i,u;\theta) \subset \partial^-L(i,u;\theta)$. By property  (P3) of the Appendix, it then follows that $\sum_{i,u}\mu(i,u)H(i,u;\theta) \subset \partial^- \bar{\E}(\theta)$. The claim follows from item 3. in (C1) given that any limit point of $\theta_n$ as $n\uparrow\infty$ must be a critical point of $\partial^-\bar{\E}(\cdot)$ in view of the foregoing. \hfill $\Box$

\bigskip

Some comments regarding  our assumptions are in order.
\begin{enumerate}

\item The vanilla Q-learning iterates, being convex combinations of previous  iterates with a bounded quantity, remain bounded. Thus the boundedness assumption on $Q$ in (C1) is reasonable. The twice continuous differentiability of $Q$ in $\theta$ is reasonable when the neural network uses a smooth nonlinearity such as SmoothReLU, GELU or a sigmoid function. As we point out later, using standard ReLU adds another layer of non-smooth analysis which we avoid here for the sake of simplicity of exposition.  The last condition in (C1) is also reasonable, e.g., when the graphs of $Q(x,u; \cdot), Q(x,u'; \cdot)$ cross along a finite union of lower dimensional submanifolds.\\

\item (C2) assumes stability of iterates, i.e., $\sup_n\|\theta_n\|  < \infty$ a.s. There is an assortment of tests to verify this. See, e.g., \cite{BorkarBook}, Chapter 3. Also, one can enforce this condition by projection onto a convenient large convex set every time the iterates exit this set, see \textit{ibid.}, Chapter 7.\\

\item (C3) entails that we perform successive experience replays over larger and larger batches of past samples so that the error in applying the strong law of large numbers decreases sufficiently fast. While this is possible in principle because of the increasing pool of past traces with time, this will be an idealization in practice. It seems possible that the additional error in absence of this can be analyzed as in \cite{Arun}.  Note also that for deterministic control problems, experience replay is not needed for our purposes. The cartpole model studied in the next section is an example of this.\\

\end{enumerate}

It is worth noting that bulk of the argument above is indeed the classical argument for convergence of stochastic gradient descent with both Markov and martingale difference noise, except that our iteration fits this paradigm only `a.s.'. The missing piece is that the (possibly random) point it converges to need not be a point of differentiability of $\bar{\E}(\cdot)$, and therefore not a classical critical point thereof. This is what calls for the back and forth between the classical proof and the differential inclusion limit for stochastic gradient descent to minimize a non-smooth objective function.

Before we proceed, we would like to underscore a subtle point, viz., the role of experience replay here. Consider the scheme without the experience replay as above, given by
\begin{eqnarray}
\theta_{n+1} &=& \theta_n - a(n)(r(X_n,U_n) + \gamma\max_v Q(X_{n+1}, v; \theta_n) - Q(X_n, U_n;\theta_n))\times \nonumber \\
&& \left(\gamma\nabla_\theta Q(X_{n+1}, v; \theta_n)\Big|_{v = \mbox{argmax} Q(X_{n+1}, \cdot ; \theta_n)} - \nabla_\theta Q(X_n, U_n; \theta_n)\right). \nonumber \\
\ && \ \label{FG-DQN1}
\end{eqnarray}
The limiting o.d.e. for this is
\begin{eqnarray}
\dot{\theta}(t) &=& E\Bigg[\sum_y p(y|X_n, U_n)\Bigg((r(X_n,U_n) + \gamma\max_v Q(y, v; \theta(t)) - Q(X_n, U_n;\theta_n))\times \nonumber \\
&& \left(\gamma\nabla_\theta Q(y, v; \theta(t))\Big|_{v = \mbox{argmax }Q(y, \cdot ; \theta(t))} - \nabla_\theta Q(X_n, U_n; \theta(t))\right)\Bigg)\Bigg], \nonumber \\
\ && \ \label{Wrongeq1}
\end{eqnarray}
where $E[ \cdot ]$ denotes the stationary expectation. This  is again not in a form where the convergence is apparent. The problem, typical of naive Bellman error gradient methods, is that we have a conditional expectation (w.r.t.\  $p(\cdot|X_n, U_n)$) of a product instead of a product of conditional expectations, as warranted by the actual Bellman error formula. The experience replay suggested above does one of the conditional expectations ahead of time, albeit approximately, and therefore renders (approximately) the expression a product of conditional expectations.
Observe that this is so because we average over past traces $(X_m, U_m, X_{m+1})$  where $X_m, U_m$ are fixed at the current $X_n, U_n$, so that it is truly a Monte Carlo evaluation of a conditional expectation. If we were to average over such traces without fixing $X_n, U_n$, we would get the o.d.e.
\begin{eqnarray}
\dot{\theta}(t) &=& E\left[r(X_n,U_n) + \gamma\max_vQ(X_{n+1}, v; \theta(t)) - Q(X_n, U_n;\theta_n)\right]\times \nonumber \\
&& E\left[\gamma\nabla_\theta Q(X_{n+1}, v; \theta(t))\Big|_{v = \mbox{argmax}Q(X_{n+1}, \cdot ; \theta(t))} - \nabla_\theta Q(X_n, U_n; \theta(t))\right], \nonumber \\
\ && \ \label{Wrongeq2}
\end{eqnarray}
where $E[ \cdot ]$ denotes the stationary expectation. Here the problem is that the desired `expectation of a product of conditional expectations' has been split into a product of expectations, which too is wrong. This discussion underscores an additional advantage of experience replay in the context of Bellman error gradient methods, over and above its traditional advantages listed earlier.

\subsection{Comments about `on-policy' schemes}

An `on-policy' scheme has an additional complication, viz., the expectation operator in the definition of $\bar{\E}( \cdot )$ itself depends on the parameter $\theta$. This is because the policy with which the state-action pairs $(X_n,U_n)$ are  being sampled depends at time $n$ on the current iterate $\theta_n$. Therefore there is explicit $\theta$ dependence for the probability  measure $\mu(\cdot, \cdot)$, now written as $\mu_\theta(\cdot, \cdot)$. The framework of \cite{Yaji} is broad enough to allow this `iterate dependence' and we get the counterpart of (\ref{incl}) with $\mu(\cdot,\cdot)$ replaced by $\mu_{\theta(t)}(\cdot,\cdot)$, leading to the limiting  differential inclusion
\begin{equation}
    \dot{\theta}(t) \in -\nabla^*\bar{\E}_{\theta(t)}(\theta(t)).
    \label{SGD2}
    \end{equation}
Here $\nabla^*$ denotes the Frechet subdifferential with respect to only the argument in parentheses, not the subscript. Hence it is not the full subdifferential and the theoretical issues we pointed out regarding DQN come back to haunt us. This is true, e.g., when you use the $\epsilon$-greedy policy that picks the control argmax$(Q(X_n, \cdot; \theta_n))$ with probability $1 - \epsilon$, and chooses a control independently and with uniform probability from $A$, with probability $\epsilon$.

Clearly, a scheme such as (\ref{SGD2}) that performs gradient descent for the stationary expectation of a parametrized cost function w.r.t.\ the parameter, but ignoring the parameter dependence of the stationary law itself on the parameter, is not guaranteed to converge. There are special situations such as the EM algorithm \cite{Delyon} where additional structure of the problem makes it work. In general, policy gradient methods based on suitable sensitivity formulas for Markov decision processes seem to provide the most flexible approach in such situations, see, e.g., \cite{Marbach}.

\subsection{Comparison with double sampling}
\label{subsec:doublesampling}

To recapitulate, DQN can be viewed as an instance of a broader class of schemes known as Bellman error minimization or Bellman residual minimization \cite{Baird}. The commonality between such schemes is that they first replace  the candidate value function by a parsimoniously  parametrized family of functions, e.g., linear combinations of  basis functions or  neural networks. The original equation then need not hold, so one seeks to minimize the `Bellman error', i.e., the squared difference between the right and left hand sides of the approximate Bellman quation. Its gradient involves a product of conditional expectations. If one uses the naive strategy of replacing them by evaluation at actual samples, the gradient of the resulting `empirical Bellman error' leads to an (approximate) expectation of a product in the averaged dynamics where it should have been the expectation of a product of conditional expectations. That is, product and conditional expectation get interchanged, causing bias to creep in. In fact, \cite{Baird} already containes a way to avoid this. This is  the  `double sampling' scheme that simulates two transitions simultaneously at each time instant for the current state-action pair. These are simulated independently with the same conditional law. One then performs the function evaluations for next state in the two terms of the product in Bellman error gradient using the two different samples thus generated. While this has been used subsequently (see, e.g., \cite{Prabu, Prabu2}),  it can be very awkward to implement in some simulation environments and is certainly untenable in on-line mode. Also, it increases the variance as we note below in numerical experiments.  One of the contributions of the present work is to circumvent this by using a variant of experience replay. This can be executed with a single simulation run with buffered data and also has the advantage of lower variance due to averaging.

As for the mathematical analysis, the error process $\{\epsilon(n)\}$ in the application of the strong law of large numbers to experience replay drops out and assumption (C3) becomes redundant if no experience replay is used. With pure double sampling without experience replay, we have only the martingale difference noise obtained by subtracting from the right hand side of the iteration its one step conditional expectation. This will be a little different from the martingale difference noise $\{M_{n+1}(\theta_n)\}$ above due to the additional simulated transition and perforce will have higher variance.

A recent work \cite{Jiang} treats the empirical Bellman error as a deterministic function of the parameter and minimizes it using the full gradient as described here. It does not, however, use either double sampling or experience replay and therefore retains the problem of replacing a product of conditional expectations by  conditional expectation of a product.

For deterministic systems, double sampling is redundant as there is no conditional expectation in the Bellman equation. Experience replay may still be desirable for its other advantages mentioned earlier, but is not required on above grounds.

\section{Numerical results}
In this section, we compare on two realistic examples the performance of FG-DQN with respect to that of the standard DQN scheme \cite{Mnih2}. In particular, we investigate the behaviour of Bellman error, Hamming distance from the optimal policy (if the optimal policy is known) and the average reward. The pseudo-code for FG-DQN is described in Algorithm \ref{fgdqn_algorithm}.

\subsection{Forest management problem}
\label{subsec:forestproblem}

Consider a Markov decision process framework for a simple forest management problem \cite{Chades14,Couture16}. The objective is to maintain an old forest for wildlife and make money by selling the cut wood. We consider discounted infinite horizon discrete-time problem. The state of the forest at time $n$ is represented by $ X_n \in \{0,1,2,3, \cdots, M\}$ where the value of the state represents the age of the forest; $0$ being the youngest and $M$ being the oldest. The forest is managed by two actions: `Wait' and `Cut'. An action is applied at each time at the beginning of the time slot. If we apply the action `Cut' at any state, the forest will return to its youngest age, i.e., state $0$. On the other hand, when the action `Wait' is applied, the forest will grow and move to the next state if no fire occurred. Otherwise, with probability $p$, the fire burns the forest after applying the `Wait' action, leaving it at its youngest age (state $0$). Note that if the forest reaches its maximum age, it will remain there unless there is a fire or action `Cut' is performed. Lastly, we only get a reward when the `Cut' action is performed. In this case, the reward is equal to the age of the forest. There is no reward for the action `Wait'.\\

\bigskip

\begin{algorithm}[H]\label{algo1}
\SetKwInOut{Input}{Input}
\SetKwInOut{Output}{Output}
\SetAlgoLined
 \textbf{Input:} replay memory $\mathcal{D}$ of size $M$, minibatch size $B$, number of episodes $N$, maximal length of an episode $T$, discount factor $\gamma$, exploration probability $\epsilon$.\\
 Initialise the weights $\theta$ randomly for the Q-Network.\\
 \For{Episode = $1$ to $N$}{
  Receive initial observation $s_1$.\\
  \For{n = $1$ to $T$}{
  \eIf{Uni[0,1] $<$ $\epsilon$ }{Select action $U_n$ at random.}
  {
   $U_n = Argmax_{u} Q(X_n,u;\theta)$
  }
  Execute the action and take a step in the RL environment.\\
  Observe the reward $R_n$ and obtain next state $X_{n+1}$.\\
  Store the tuple $(X_n, U_n, R_n, X_{n+1})$ in $\mathcal{D}$.\\
  Sample random minibatch of $B$ tuples from $\mathcal{D}$.\\
  \For{ k =  $1$ to $B$}{
    Sample all tuples $(X_j, U_j, R_j, X_{j+1})$ with a fix state-action pair $(X_j = X_k, U_j = U_k)$ from $\mathcal{D}$
    \[Set\, Z_j = \begin{cases} R_j, & \mbox{for terminal state,} \\
    R_j + \gamma \max_{u} Q(X_{j+1},u;\theta), & \mbox{otherwise.}\end{cases}\]\\
  Compute gradients and using Eq. \eqref{FG-DQN_expreplay}
  update parameters $\theta$.\\
  }
  }
 }
\caption{Full Gradient DQN (FG-DQN)}
\label{fgdqn_algorithm}
\end{algorithm}

\bigskip

Since the objective is to maximize the discounted profit obtained by selling wood, we may want to keep waiting to get the maximum possible reward, but there is an increasing chance that the forest will get burned down.\\

\begin{figure}%
\centering
\subfigure[Average loss and 95\% confidence interval]{%
\label{fig1a}%
\includegraphics[scale = 0.75]{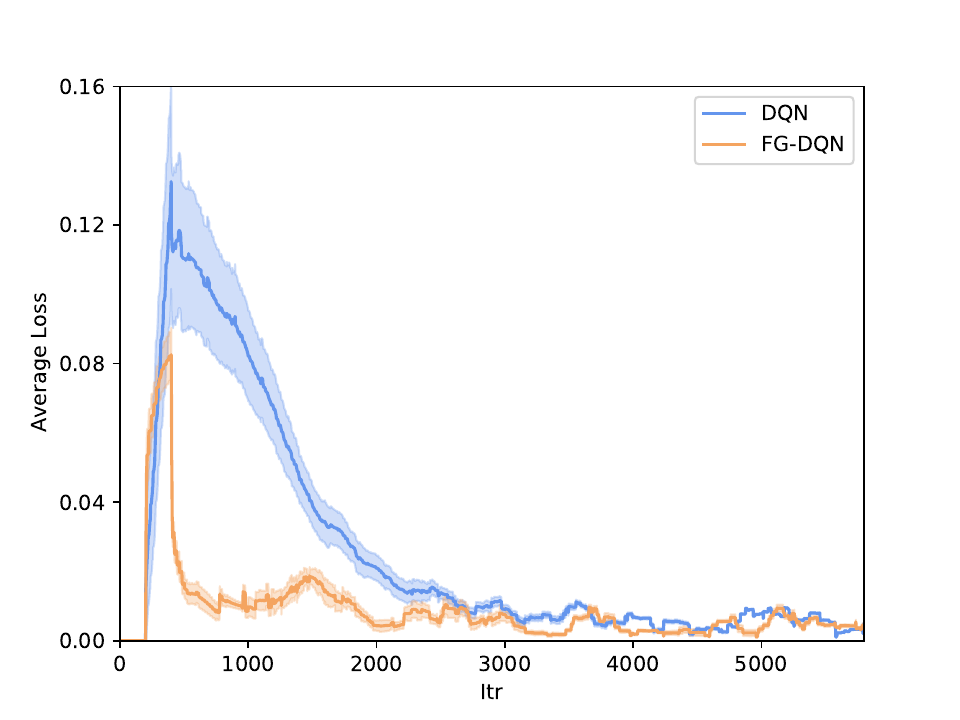}}%

\subfigure[Average Hamming distance from the optimal policy]{%
\label{fig1b}%
\includegraphics[scale = 0.75]{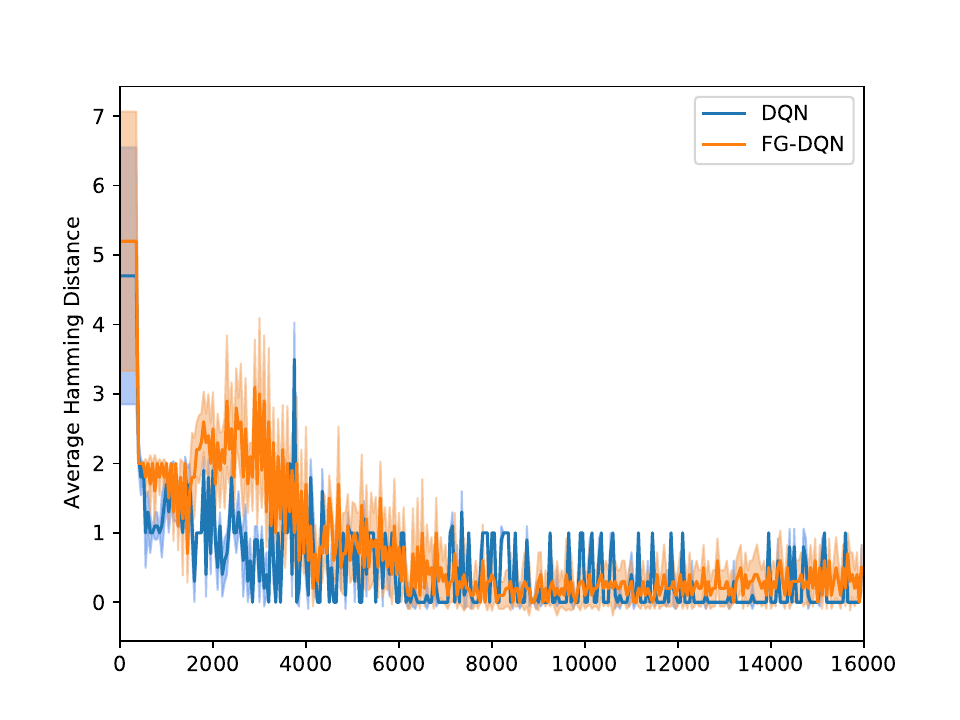}}%
\caption{Forest management problem with $\gamma = 0.8$ and $p = 0.05$}
\label{fig1}
\end{figure}

\begin{figure}%
\centering
\subfigure[Average loss and 95\% confidence interval]{%
\label{fig2a}%
\includegraphics[scale = 0.71]{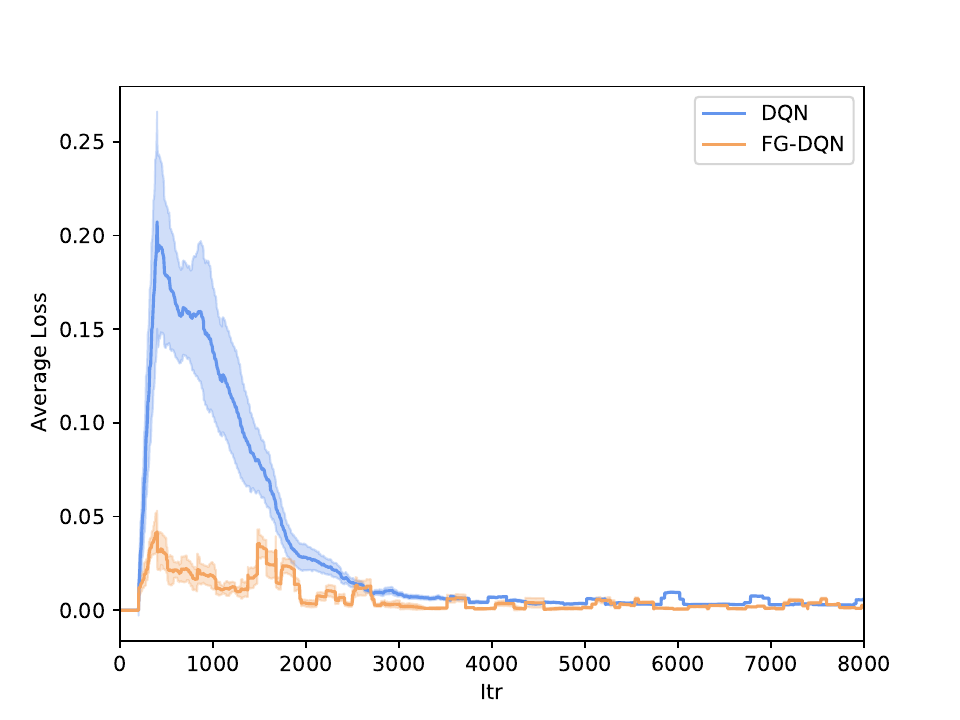}}%
\hspace{-1em}
\subfigure[Average Hamming distance from the optimal policy]{%
\label{fig2b}%
\includegraphics[scale = 0.71]{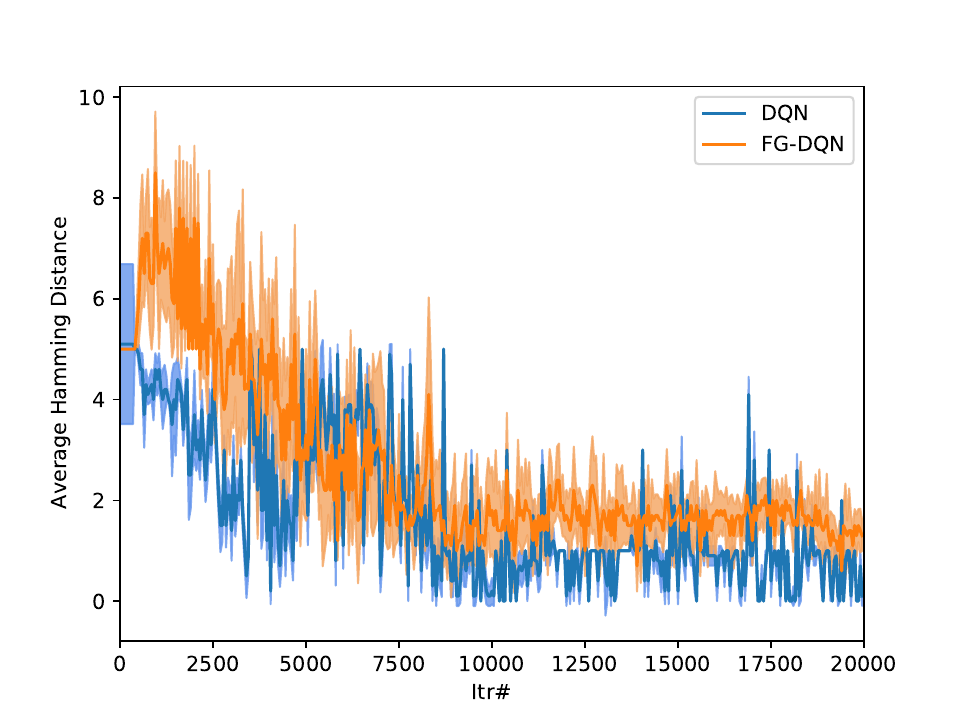}}%
\caption{Forest management problem with $\gamma = 0.95$ and $p = 0.01$}
\end{figure}

For numerical simulations, we assume that the maximum age of the forest is $M = 10$. Then, we implement standard DQN and FG-DQN to analyse the policy obtained from the algorithm and the Bellman error. We use a neural network with one hidden layer to approximate the Q-value. The number of neurons for this hidden layer is  $2000$, and we use ReLU for nonlinear activation. It has been recently advocated to use a neural network with one but very wide hidden layer \cite{Agazzi21,Chizat18}. The input to the neural network is the state of the forest and the action. Furthermore, the batch size to draw the samples for the experience replay is fixed to $25$. We test both the algorithms for the off-policy scheme, i.e., we run through all possible state-action pairs in round-robin fashion to train the neural network.

We run two different simulations - i) with low discounting factor $\gamma = 0.8$ and ii) with high discounting factor $\gamma = 0.95$. Fig.~\ref{fig1} depicts the simulation results for case i) with forest fire probability $p = 0.05$.  We run the experiment $10$ times and plot the running average of Bellman error across iterations in Fig.~\ref{fig1a}. We also calculate the standard deviation of the Bellman error.  The shaded region in the plot denotes the $95\%$ confidence interval. We observe that FG-DQN converges much faster than DQN. Furthermore, the variance for FG-DQN is relatively low.

We now analyse how far the answer of each algorithm is from the optimal policy. To do this, we first find the optimal policy for this setting by policy iteration algorithm. The optimal policy has a threshold structure as follows: $\pi^* = [0,0,1,1,1,1,1,1,1,1]$
for $\gamma=0.8$ and $p=0.05$.

After each iteration, we now evaluate the Q-network and calculate the Hamming distance between the current policy and the optimal policy  $\pi^*$, which gives us the count of the number of states where optimal action is not taken.  We run the simulations $10$ times and plot the average Hamming distance for DQN and FG-DQN in Fig.~\ref{fig1b}. Note that we plot every $50th$ value of the average Hamming distance in the figure. It is to avoid the squeezing of rare spikes obtained at later time steps of the simulations.
The shaded region denotes the $95\%$ confidence interval for the averaged Hamming distance. We observe from the figure that the policy obtained by FG-DQN starts converging to the optimal policy at around $8000$ iterations. In comparison, for DQN, we observe a lot of spikes during later iterations. The occurrence of these spikes means that there is a one-bit error in the policy obtained by DQN. Further analysis shows that the DQN policy in this case which has one bit error resembles to myopic policy $[0,1,1,1,1,1,1,1,1,1]$.

We next observe the impact of a high discounting factor on the performance of our algorithm and how well it performs as compared to the standard DQN scheme. We set $\gamma = 0.95$ and forest fire probability $p=0.01$. The optimal policy obtained by exact policy iteration for this case is $\pi^* = [0,0,0,0,0,1,1,1,1,1]$.  Fig~\ref{fig2a} shows the mean loss for $10$ simulations and the corresponding $95\%$ confidence interval. We observe similar behaviour as before, i.e.,  the variance for FG-DQN is low. Fig~\ref{fig2b} shows the averaged Hamming distance between the policy obtained by the algorithm and the optimal policy. It is clear from the figure that the variance for DQN is very high throughout the simulation. It means we may end up with a policy that can have a $3$ or $4$  bits error at the end of our simulation runs. On the other hand, FG-DQN is more stable since it shows fewer variations with the increasing number of iterations. Thus, we are more likely to get the policy with a $2$ bits error on average. The shaded region in the plot shows the $95\%$ confidence interval for $10$ simulations which demonstrates that the behaviour is consistent across the multiple simulations.

\begin{figure}%
\centering
\includegraphics[scale = 0.71]{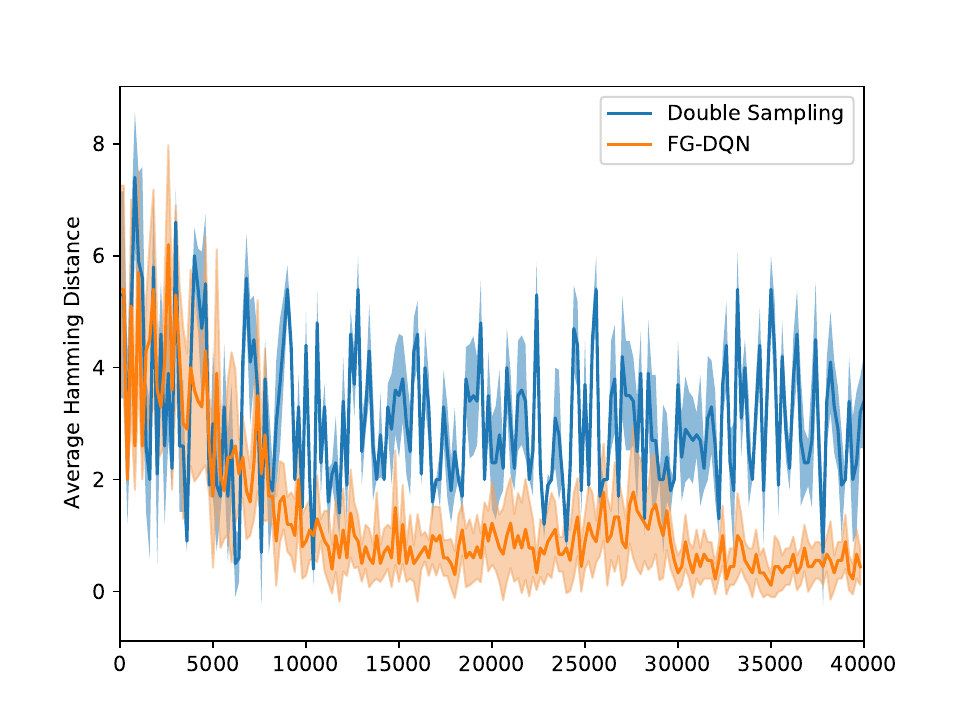}%
\caption{Comparison with the double sampling scheme. Average Hamming distance from the optimal policy for the forest management problem with resetting to a uniform value and with $\gamma = 0.9$ and $p = 0.2$.}
\label{Fig_stoch_DS}
\end{figure}

Let us present an additional simulation to evaluate the performance of FG-DQN versus double sampling scheme \cite{Baird}. We note that the double sampling scheme requires to generate two independent samples at each time step. This becomes difficult in many simulation environments and impossible in on-policy mode. We further note that if the underlying environment is deterministic, both these schemes become exactly identical. Therefore, in order to investigate the difference in their performance,  we slightly modify the forest management problem to have more stochasticity in its dynamics.  Namely, the dynamics remain the same except for the following change.   With probability $p$, the fire burns
a fraction of the forest after applying the `Wait' action. The fraction of the forest burnt follows a uniform distribution. In this simulation, we set $p=0.2$ and the discount factor $\gamma = 0.9$.  The optimal policy obtained by the exact policy iteration for this case is $\pi^* = [0,0,1,1,1,1,1,1,1,1]$. Fig.~\ref{Fig_stoch_DS} shows the comparison of averaged Hamming distance between the policy obtained by respective algorithm and the optimal policy. Note that we run the simulations $10$ times and also plot the $95\%$ confidence intervals. We observe that the policy obtained from FG-DQN approaches quicker the optimal policy and the performance is more stable.  On the other hand, the double sampling policy has significant fluctuations even after $30000$ iterations.  The figure also shows that the double sampling policy has a $2-4$  bits error at the end of our simulation runs.

\subsection{Cartpole - OpenAI Gym model}

\begin{figure}%
\centering%
\includegraphics[scale = 0.81]{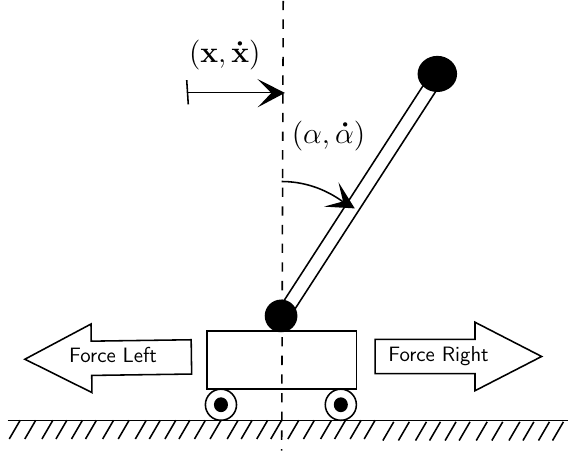}
\caption{Cartpole system}
\label{figcartpole}
\end{figure}

We now test our algorithm for a more complex example, the Cartpole-v0 model from OpenAI gym \cite{brockman2016}.  The environment description is as follows. The state of the system is defined by a four dimensional tuple that represents cart position $x$, cart velocity $\dot{x}$, pole angle $\alpha$ and angular velocity $\dot{\alpha}$ (See Fig.~\ref{figcartpole}). The pole starts upright and the aim is to prevent it from falling over by pushing the cart
to the left or to the right (binary action space).
The cart moves without friction along the $x$-axis.

We run multiple simulations, each with $1500$ episodes for DQN and FG-DQN.
For every time-step while an episode is running, we get the reward of $+1$.
The episode ends if any of the following conditions holds: the pole is more than $12^\circ$ degrees from the vertical axis, the cart moves more than $2.4$ units from the centre, or the episode length is more than $200$. The model is considered to be trained well when the discounted reward
is greater than or equal to
195.0 over 100 consecutive trials.

In this example, we used the `on-policy' version with the popular `$\epsilon$-greedy' scheme which picks the current guess for the optimal (i.e., the control that maximizes $Q(X_n, \cdot ;  \theta_n)$) with probability $1 - \epsilon$ and chooses a control uniformly with probability $\epsilon$ for a prescribed $\epsilon > 0$. We use $\epsilon = 0.1$. As we see below, FG-DQN continues to do much better than DQN even in this on-policy scheme for which we do not have a convergence proof as yet.

We use a neural network with three hidden layers. The number of nodes for the hidden layers are $16,32,$ and $32$, respectively. For non-linearity, we use ReLU activation after each hidden layer.

We now compare the performances of FG-DQN and DQN for a very high discounting factor of $0.99$. Note that the Cartpole example is deterministic, meaning that for a fixed state-action pair $(X_n, U_n)$, the pole moves to state $X_n'$ with probability $1$. As a result, there will be no averaging in eq.~\eqref{FG-DQN} and no need for `experience replay'. Since this example is complex with significant non-linearity, we use the batch size of $128$ for both DQN and FG-DQN  to update the parameters of the neural network inside one iteration.

\begin{figure}%
\centering
\subfigure[Average rewards for a typical single simulation run]{%
\label{fig3a}%
\includegraphics[scale = 0.71]{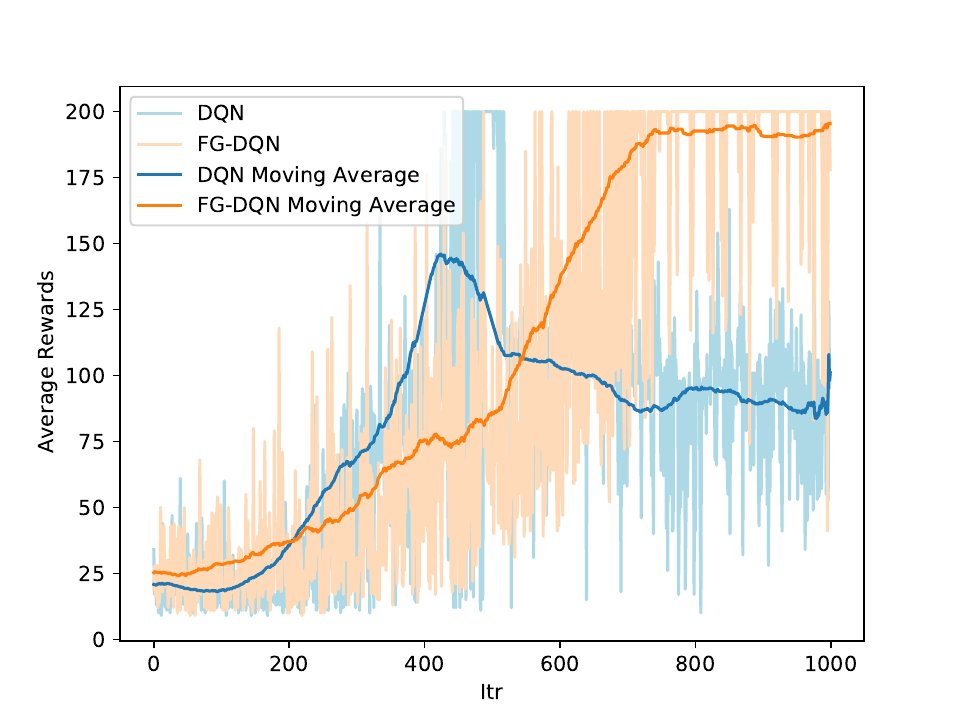}}%
\hspace{-1em}
\subfigure[Rewards averaged over simulations for Cartpole and 95\% confidence intervals]{%
\label{fig3b}%
\includegraphics[scale = 0.71]{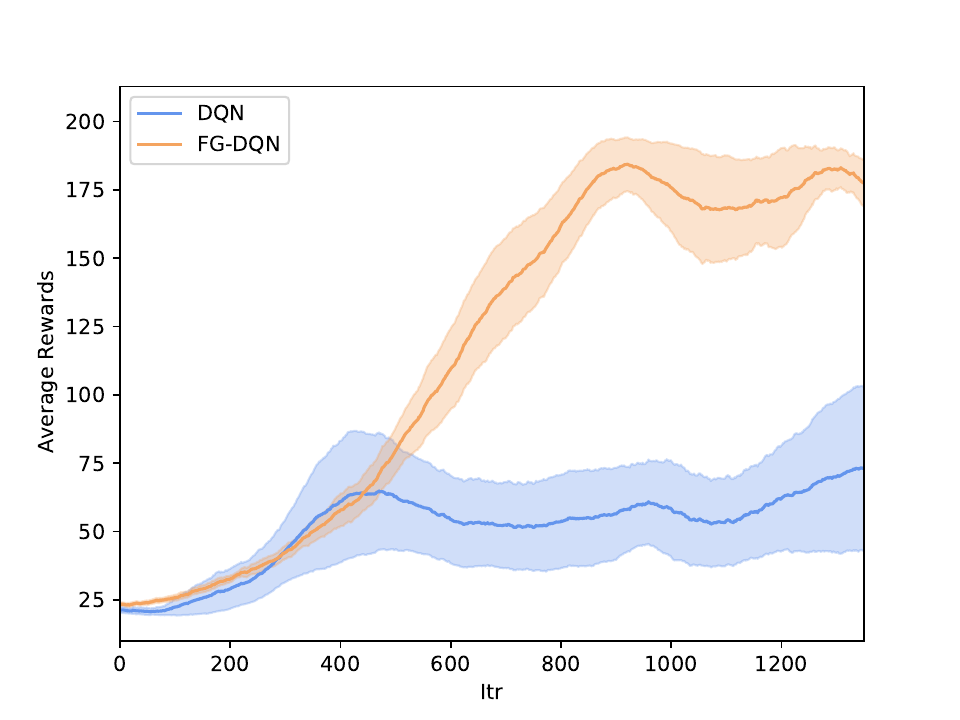}}%
\caption{Cartpole example with $\gamma = 0.99$}
\end{figure}

Fig.\ref{fig3a} depicts the reward behaviour for a single typical run of DQN and FG-DQN. We see that the fluctuations for reward per episode for both the algorithms are high, and thus, we also plot the moving average of rewards with a window of $100$ episodes. It is clear from the figure that FG-DQN starts achieving the maximum reward of $200$ after $800$ episodes regularly, however, DQN hardly attains the maximum reward during $1000$ episodes. To check the consistency of the behaviour of our algorithm, we run the experiment $10$ times and plot the average reward and $95\%$ confidence interval in Fig. \ref{fig3b}. We see that FG-DQN performs much better than DQN with an average reward after $1000$ episodes lying around $175$. In comparison, the average reward for DQN is between $50$ and $75$.

\section{Conclusions and future directions}

We proposed and analyzed a variant of the popular DQN algorithm that we call Full Gradient DQN or FG-DQN wherein we also include the parametric gradient (in a generalized sense) of the target. This leads to a provably convergent scheme with sound theoretical basis which also shows improved performance over test cases. There is ample opportunity for further research in this direction, both theoretically and in terms of actual implementations.
To highlight opportunities, we state here some additional remarks, which also contain a few pointers to future research directions.

\begin{enumerate}

\item Since the critical points are isolated, we get a.s.\ convergence to a single sample path dependent critical point. This situation is generic in the sense that  it holds true for the problem parameters in an open dense set thereof, by a standard fact from Morse theory in the smooth case. However, connected sets of non-isolated equilibria can occur due to overparametrization and it will be interesting to develop sufficient conditions for point convergence.

\item Thanks to the addition of $\{\xi_n\}$, the noise in FG-DQN is `rich enough' in all directions in a certain sense. One then expects it to ensure that under reasonable assumptions, the unstable equilibria, here the critical points other than  local minima of the Bellman error, will be avoided with probability one. That is, a.s.\ convergence to a local minimum can be claimed. See section 4.3 of \cite{BorkarBook} for a result of this flavor under suitable technical conditions. We expect a similar result to hold here.  In practice, the extraneous noise $\{\xi_n\}$ is usually unnecessary and the inherent numerical errors and noise of the iterations suffice.

\item We can also use approximation of the `max' operator by `softmax', i.e., by picking the control with a probability distribution that concentrates on argmax and depends smoothly on the parameter $\theta$. Then we can work with a legitimate gradient in place of a set-valued map in the o.d.e.\ limit, at the expense of picking up an additional  bounded error term. Then the convergence  to a small neighborhood of  an equilibrium may be expected, the size of which will be dictated in turn by the bound on this error, see, e.g., \cite{Arun}. There is a similar issue if we drop (C3) and let a persistent small error due to the use of approximate conditional expectation by experience replay remain.

\item Working with nondifferentiable nonlinearities such as ReLU raises further technical issues in analysis that need to be explored. This will require further use of non-smooth analysis.

\item As we have pointed out while describing our numerical experiments on the cartpole example, FG-DQN gives a significantly better performance than DQN, in an `on-policy' scenario for which we do not have rigorous theory yet. This is another promising and important research direction for the future.
\end{enumerate}

\section{Acknowledgement} \label{apis99}

The authors are greatly obliged to Prof.\ K.\ S.\ Mallikarjuna Rao for pointers to the relevant literature on non-smooth analysis. The work of VSB was supported in part by an S.\ S.\ Bhatnagar Fellowship from the Council of Scientific and Industrial Research, Government of India.
The work of KP and KA is partly supported by ANSWER project PIA FSN2 (P15 9564-266178 / DOS0060094) and the project of Inria - Nokia Bell Labs ``Distributed Learning and Control for Network Analysis''. This work is also partly supported by the project IFC/DST-Inria-2016-01/448 ``Machine Learning for Network Analytics''.

This is the author's version of the work for the Springer book:
A.B. Piunovskiy and Y. Zhang  (eds.),
{\it Modern Trends in Controlled Stochastic Processes: Theory and Applications,
Volume III}, 2021.

\bigskip

\noindent \textbf{\large Appendix: Elements of non-smooth analysis}\label{apiappendix9}

\bigskip

The (Frechet) sub/super-differentials of a map $f: \R^d \mapsto \R$ are defined by
\begin{eqnarray*}
\partial^-f(x) &:=& \left\{z \in \R^d: \liminf_{y\to x}\frac{f(y) - f(x) - \langle z, y-x\rangle}{|x-y|} \geq 0\right\}, \\
\partial^+f(x) &:=& \left\{z \in \R^d: \limsup_{y\to x}\frac{f(y) - f(x) - \langle z, y-x\rangle}{|x-y|} \leq 0\right\},
\end{eqnarray*}
respectively. Assume $f, g$ is Lipschitz. Some of the properties of $\partial^{\pm}f$ are as follows.
\begin{itemize}

\item \textbf{(P1)} Both $\partial^-f(x), \partial^+f(x)$ are closed convex and  are nonempty on dense sets.

\item \textbf{(P2)} If $f$ is differentiable at $x$, both equal the singleton $\{\nabla f(x)\}$. Conversely, if both are nonempty at $x$, $f$ is differentiable at $x$ and they equal $\{\nabla f(x)\}$.

\item \textbf{(P3)} $ \partial^-f +  \partial^-g   \subset  \partial^-(f + g), \  \partial^+f +  \partial^+g  \subset \partial^+(f + g) .$

\end{itemize}

The first two are proved in \cite{Bardi}, pp.\ 30-1. The third follows from the definition.
Next consider  a continuous function $f: \R^d\times B \mapsto \R$ where $B$ is a compact metric space. Suppose $f( \cdot , y)$ is continuously differentiable uniformly w.r.t.\ $y$. Let $\nabla_xf(x,y)$ denote the gradient of $f(\cdot, y)$ at $x$.  Let $g(x) := \max_yf(x,y), h(x) := \min_yf(x,y)$ with
$$
M(x) := \{\nabla_xf(x, y), y \in \mbox{Argmax} f(x, \cdot)\}
$$
and
$$
N(x) := \{\nabla_xf(x,y), y \in \mbox{Argmin} f(x, \cdot)\}.
$$
Then $N(x), M(x)$ are compact nonempty subsets of $B$ which are upper semi-continuous in $x$ as set-valued maps. We then have the following general version of Danskin's theorem \cite{Danskin}:

\begin{itemize}

\item \textbf{(P4)} $\partial^-g(x) = \overline{\mbox{co}}(M(x)), \partial^+g(x) = y$ if $M(x) = \{y\}$, $= \phi$ otherwise, and
$g$ has a directional derivative in any direction $z$ given by $\max_{y\in M(x)}\langle y, z\rangle$.\\

\item \textbf{(P5)} $\partial^+h(x) = \overline{\mbox{co}}(N(x)), \partial^-h(x) = y$ if $N(x) = \{y\}$, $= \phi$ otherwise, and
$h$ has a directional derivative in any direction $z$ given by $\min_{y\in N(x)}\langle y, z\rangle$.

\end{itemize}

The latter is proved in \cite{Bardi}, pp.\ 44-6, the former follows by a symmetric argument.


\end{document}